# Temporal Pattern Mining for Analysis of Longitudinal Clinical Data: Identifying Risk Factors for Alzheimer's Disease

Annette Spooner[1]*, Gelareh Mohammadi[1], Perminder S. Sachdev[2], Henry Brodaty[2], Arcot Sowmya[1]

[1] School of Computer Science and Engineering, UNSW Sydney, Sydney, Australia.
[2] Centre for Healthy Brain Ageing (CHeBA), Discipline of Psychiatry and Mental Health, Faculty of Medicine and Health, UNSW Sydney, Sydney, Australia.

***Corresponding author:** Annette Spooner (a.spooner@unsw.edu.au)

## Abstract

A novel framework is proposed for handling the complex task of modelling and analysis of longitudinal, multivariate, heterogeneous clinical data. This method uses temporal abstraction to convert the data into a more appropriate form for modelling, temporal pattern mining, to discover patterns in the complex, longitudinal data and machine learning models of survival analysis to select the discovered patterns.

The method is applied to two real-world studies of Alzheimer's disease (AD), a progressive neurodegenerative disease that has no cure. The patterns discovered were predictive of AD in survival analysis models with a Concordance index of up to 0.8. This is the first work that performs survival analysis of AD data using temporal data collections for AD. A visualisation module also provides a clear picture of the discovered patterns for ease of interpretability.

**Keywords:** temporal pattern mining, temporal abstraction, clinical data, survival analysis, relative risk.

## 1. Introduction

Modelling of multivariate data over time is a complex task, even more so when that data are clinical data. Methods that are typically used to analyse multivariate time-series data, such as dynamic time warping [1] or recurrent neural networks [2], rely on having frequent, equally spaced points in time. Clinical data, however, are often collected in waves, measuring various patient attributes at often infrequent and sometimes irregular intervals over time, that can span days, months or even years. Therefore, the number of data points may be quite small and not always equally spaced. Values can be missing, the data are often censored, are generally of mixed types and may not be normally distributed. In addition, any analysis done on clinical data must be interpretable by clinicians if they are to have confidence in the results. Therefore, new methods are needed to analyse this type of data.

Pattern mining, a technique used to discover relationships amongst items in a database, is one technique that has merit in overcoming these challenges. Although frequent pattern mining is a well-established and mature area of research, few publications in this area focus on the application of pattern mining to clinical data [3], because of the additional complexities this entails.

In frequent pattern mining, the items to be mined are unordered and have no time dimension. Temporal pattern mining, on the other hand, mines sequences of intervals. Since the items to be mined span an interval of time rather than just a single time point, the ordering of the items is important, and there are complex relationships amongst these intervals [4]. When applied to clinical data, the temporal patterns identified should be clinically relevant. In the case of rare or relatively uncommon diseases, any patterns associated with disease progression are unlikely to occur frequently in the overall population, as the number of individuals suffering from the disease is small [3]. Therefore, a method for mining clinical data must combine techniques for temporal pattern mining, rare pattern mining and temporal abstraction, as a means of summarising the state of a patient over a time.

To address these challenges, we propose a new temporal pattern mining framework for clinical data that identifies high risk patterns and presents clearly interpretable results for clinicians. We name this method Clinical Temporal Pattern Mining or C-TPM. C-TPM is able to identify patterns that are rare overall but occur frequently in the at-risk population. It does this using the concepts of relative risk and the odds ratio to identify interesting and high-risk patterns in the data [3]. Both of these concepts are widely used in epidemiological studies.

The proposed method is based on that of Batal et al. [5], with some modifications, and involves four steps:
1. Temporal abstraction [6] is used to create a higher-level view of the clinical data, allowing heterogeneous data to be represented similarly for comparison.
2. A temporal pattern mining algorithm, based on TPMiner [7], is used to mine for high-risk patterns from the abstracted clinical data.
3. The patterns identified are converted into a binary matrix, with one row per patient and one column per pattern. A '1' in a row/column indicates the patient's data contain that pattern and a '0' indicates otherwise. This matrix is used as input to a machine learning survival analysis model to determine the patterns that are predictive.
4. A visualisation module displays the most predictive patterns in a clearly interpretable manner.

The temporal patterns discovered by this method consist of a sequence of one or more patient states that hold for an interval of time and the relationships between those intervals. A patient state consists of a feature and its abstracted value, such as <blood pressure – high>. For simplicity. the 13 possible relations between intervals identified by Allen [4] have been simplified to two - just co-occurs and follows - after the method of Batal et al [5]. An example of a temporal pattern might be <blood pressure – high> for 2 waves, followed by <cholesterol – high> for 1 wave. Another example could be <lateral stability – poor> for 2 waves, co-occurring with <chemical exposure – high> for 2 waves.

To demonstrate the applicability of this method, it has been applied to two real-world studies of Alzheimer's disease (AD). AD is an irreversible neurodegenerative disease affecting cognitive function.  It is a relatively uncommon disease, affecting 1 in 9 people (11.3%) over the age of 65 in the United States [8]. The underlying pathological processes leading to this disease begin at least 2-3 decades before overt symptoms appear [9]. Modelling the progression of this disease over time may lead to a better understanding of the biological processes behind it and could help in the early detection and management of the disease.

The contributions of this work are multiple. We present a novel framework that is capable of modelling complex, multivariate longitudinal clinical data and discovering patterns that are highly predictive of AD. This framework extends the work done by Batal et al [5] and Kocheturov et al

[10] to mine relevant patterns from clinical data by using a more efficient pattern mining algorithm. It also extends the TPMiner algorithm [7] to work with clinical data that are collected in waves. It applies temporal pattern mining to censored data and to the task of AD prediction. It applies temporal abstraction to heterogeneous and discrete-valued data, as well as continuous numeric data, and identifies a complete set of cut-off values for discretisation and interpretation of this data that is applicable to studies on ageing populations. These cut-off values can be found in the supplementary material.

## 2. Related Work

Frequent pattern mining was first introduced by Agrawal and Srikant [11], who developed the Apriori algorithm to identify items in a dataset that often occur together. Its main use was in market basket analysis, to understand buying patterns and predict customer behaviour. Apriori is an iterative algorithm that at each step generates new candidates using those from the previous step, counts the *support* of each candidate (the number of times it appears in the database) and prunes those candidates that do not meet a minimum support threshold (*minsup*).

The Apriori algorithm is effective, but inefficient for several reasons: (i) it may generate candidates that do not exist in the database and waste time considering these patterns, (ii) it scans the database on each iteration to calculate candidate support and (iii) it can consume large amounts of memory, as it stores all candidates in memory [12].

Many improvements to the Apriori algorithm have been proposed since it was first introduced. The Eclat algorithm converts the database into a vertical format, by associating with each item a list of transactions in which it occurs [13]. This method requires only a small number of database scans, but still requires candidate generation.

A major advance was the FP-Growth algorithm, which requires no candidate generation and only two scans of the database [14]. The FP-Growth algorithm initially scans the database to count the support of each item, as in Apriori, but on the second scan it constructs a data structure, known as the frequent pattern tree (FP-tree), which represents the database in a compact form and, with the addition of header tables, allows all frequent patterns to be found by scanning the FP-tree.

Sequential data contain sets of instantaneous time points, such as the price of shares on the stock market at any given time. Unlike frequent itemset mining, where the ordering of items within a set is irrelevant, the ordering of sequential items is important. The FP-Growth algorithm cannot be applied to sequential data because it relies on re-ordering the items when creating the FP-tree, to facilitate efficient mining of items from the tree. When mining sequential data, the original ordering of the items must be preserved, and so the FP-tree structure loses its power [15].

To overcome this problem the PrefixSpan algorithm was developed to mine sequential data [15]. PrefixSpan, a divide and conquer strategy, is based on the FP-Growth algorithm. At each step the algorithm recursively "projects" a sequence database into a set of smaller databases and then mines frequent patterns in each smaller projected database. Because the projected databases become progressively smaller, the pattern mining becomes more efficient at each step.

Temporal data adds yet another level of complexity to pattern mining, because the items being mined are intervals of time, rather that instantaneous time points. Not only is the ordering of items important, but the duration of the intervals matters as well. Several algorithms have been developed to mine for temporal patterns [16] [17] [18] [19] [20] [21] [7], but most are based on the Apriori algorithm which, as already explained, is inefficient. Only two algorithms are based on

PrefixSpan – TPrefixSpan [18] and TPMiner [7]. TPMiner introduces a new, unambiguous, endpoint representation to define the temporal intervals and some advanced pruning strategies to improve efficiency.

When modelling a disease such as AD that is relatively uncommon, any patterns associated with disease progression are unlikely to occur frequently in the overall population, as the number of patients suffering from the disease is small. Therefore, the pattern mining technique must be able to identify rare patterns. The simplest way to search for rare patterns is to simply lower the minimum support threshold. However, this can generate an enormous number of patterns, many of which will not be clinically relevant, and will also increase the computational complexity of the algorithm, to the point where it is likely to become computationally infeasible. So other methods must be considered.

To minimise the number of spurious patterns discovered, it is important to choose an appropriate 'interestingness measure' to identify relevant patterns. Interestingness measures can be used to improve mining efficiency by pruning uninteresting patterns during the mining process, thus narrowing the search space [22]. The minimum support threshold in a frequent pattern mining algorithm is an example of this. Interestingness measures are also used to select and rank patterns according to their potential interest [22].

The concepts of relative risk and the odds ratio are widely used in epidemiological studies and have recently been used as interestingness measures in pattern mining [3] [23] [24]. Relative risk is a ratio of the probability of an event occurring in the exposed group (here the group with the pattern) versus the probability of the event occurring in the non-exposed group (here the group without the pattern) [25]. It therefore indicates the increased or decreased likelihood of an event i.e. the diagnosis of AD, based on the presence of a given pattern in the patient's data.

The odds ratio is the odds of the event in the exposure group divided by the odds of the event in the control or non-exposure group [25] and also indicates an increased or decreased likelihood of an event. While the relative risk and odds ratio are similar, they are not the same. The relative risk is a ratio of probabilities while the odds ratio is a ratio of odds, which is itself a ratio of the number of events versus the number of non-events. The interested reader may consult the references for further information.

Use of the relative risk or the odds ratio as interestingness measures when mining clinical data allows the minimum support threshold to be lowered so that less common patterns can be identified without increasing the number of spurious patterns found to an unmanageable level.

To facilitate the mining of clinical data, Shahar and Musen [6] developed the concept of knowledge-based temporal abstraction, the goal of which is to evaluate and summarise the state of a patient over a period of time. Temporal abstraction:
1) summarises and creates a higher-level view of the data that is less noisy
2) represents heterogeneous variables similarly for comparison
3) helps detect trends and patterns in temporal data
4) handles missing values and asynchronous data and
5) improves interpretability.

Orphanou et al. [26] showed that temporal abstraction can yield better performance in medical contexts where the disease develops over a long period of time. Temporal abstraction relies on discretising the data in a meaningful way. Maslove et al. [27] evaluated six discretisation strategies for clinical data, both supervised and unsupervised, as well as cut-off values (RR), sets of values used to interpret medical test results, that are determined from a variety of online sources. They pointed out that the partitions resulting from the discretisation of clinical data should reflect the original distribution of the continuous attribute, maintain any patterns in the attribute without adding spurious ones, and be interpretable and meaningful to domain experts. They found that supervised discretisation tended to produce more accurate classifiers than unsupervised, and that cut-off values, where available, were effective with clinical labels, but less so with cluster labels.

Batal et al. [21] combined temporal abstraction with temporal pattern mining to classify multivariate time series. They first transformed the data using temporal abstraction, then mined temporal patterns from the data using an efficient pruning strategy. Finally, they transformed the patterns found by the algorithm into a binary matrix, with one row per patient and one column per pattern. A '1' in a particular row and column indicates that that pattern was found in the data for that patient, a '0' indicates otherwise. This binary matrix was then analysed by standard machine learning classification algorithms.

Batal et al.'s algorithm is based on the Apriori algorithm, with an advanced pruning strategy, called the minimal predictive temporal patterns (MPTP) framework, to filter out non-predictive and spurious temporal patterns. The method was tested on a dataset containing only five features – one Boolean feature and four continuous numeric features. So, it does not address the issues of heterogeneous data or higher-dimensional datasets, nor is the Apriori algorithm scalable.

Kocheturov et al. [10] extended Batal's work by transforming the abstracted data into a vertical list format prior to mining. Their method produced a significant speedup in computational time, but at the expense of memory usage. The work presented here extends Batal's work even further, mining temporal patterns with an FP-Growth-based algorithm which is more efficient both computationally and in its usage of memory, enabling it to handle a far larger number of features. The method is adapted to survival analysis and a complete set of cut-off values for AD data are provided.

## 3. Methods

### 3.1 Data

Experiments in this work were conducted using data from two studies of older Australians, the Sydney Memory and Ageing Study (MAS) [28] and the Older Australian Twin study (OATS) [29]. The characteristics of both studies are summarised in Table 1Table 1.

|  | MAS | OATS |
| --- | --- | --- |
| **Study design** | Longitudinal cohort study | Longitudinal twin study |
| **Age at baseline** | 70-90 years | 65 years |
| **Sample size (n)** | 1037 | 623 |
| **Number of clinical variables** | 70 | 44 |
| **Number of waves of data** | 7 | 3 |
| **Interval between waves** | 2 years | 2 years |
| **Number of AD diagnoses** | 146 | 26 |
| **Censoring rate** | 86% | 96% |

*Table 1. Study characteristics*

### 3.1.1 Sydney Memory and Ageing Study (MAS)

MAS is a longitudinal population-based cohort study aimed at examining the characteristics and prevalence of mild cognitive impairment and dementia. MAS began in 2005 with 1037 community-dwelling adults, aged 70-90 years, who were randomly recruited from the electoral roll of two federal government electoral areas in eastern Sydney, Australia. Data were collected every two years, culminating in seven waves of data. Full details of the study can be found [28].

The MAS dataset contains a diverse collection of data including demographics, genetics, objective cognitive data, subjective cognitive ratings, medical history, family history, medical examination, psychological scores, functional data and quality of life ratings [28]. The event of interest in the survival analysis was a diagnosis of possible or probable Alzheimer's disease, the most common form of dementia, over a period of 12 years, from waves 1 to 7 of the study. During this period 146 people developed Alzheimer's disease, indicating a censoring rate of 86%.

The Human Research Ethics Committees of the University of New South Wales and the South Eastern Sydney and Illawarra Area Health Service granted ethics approval for the MAS and OATS studies and written informed consent was given by all participants and informants. Both studies and this work were carried out in accordance with the relevant Governance guidelines, which are based on relevant University of New South Wales and National Health and Medical Research Council research and ethics policies.

### 3.1.2 Older Australian Twin Study (OATS)

OATS is a longitudinal study of twins, aged 65 years or older. Its aim is to investigate genetic and environmental factors and their interactions in healthy brain ageing and neurocognitive disorders. The initial cohort comprised 623 individuals (161 monozygotic and 124 dizygotic twin pairs; 1 set of monozygotic triplets; 27 single twins and 23 non-twin siblings). Because this study uses pattern mining, which is not affected by correlated data, all of the available data were utilised. Full details of the study can be found [29].

The data collected by the OATS study are similar to that collected by MAS and include comprehensive psychiatric, neuropsychological, cardiovascular, metabolic and neuroimaging assessments at two-yearly intervals. The event of interest in the survival analysis was a diagnosis of possible or probable Alzheimer's disease, over a period of 4 years, from waves 1 to 3 of the study. During this period 26 people developed Alzheimer's disease, indicating a censoring rate of 86%.

Participants provided written informed consent. The study was approved by the ethics committees of the Australian Twin Registry, University of New South Wales, University of Melbourne, Queensland Institute of Medical Research and the South Eastern Sydney & Illawarra Area Health Service.

### 3.2 Methodology

The methodology used in this work is shown in Figure 1 and each of the steps shown is described in more detail in Sections 3.2.1 to 3.2.5.

### 3.2.1 Temporal Abstraction

Temporal abstraction is the task of creating higher-level temporal concepts and patterns from raw time-stamped data [30]. The first step in temporal abstraction is discretising the data in a meaningful way. Some clinical variables have a well-accepted set of reference values that divide the population into groups. One example of this, shown in Table 2, is the body mass index (BMI), which provides an estimate of body fat. Where available, cut-off values have been used to discretise the data and have been determined from a variety of online sources. A complete list of these can be found in the supplementary material.

Many features, especially composite scores that are created from multiple raw features, do not have established reference levels. Batal et al. [21] used percentiles (5th, 25th, 75th and 95th percentile), or equal frequency interval binning [27], to discretise their continuous data, as shown in Table 3. We have followed this method for continuous-valued features.

For discrete-valued features and composite scores, levels can collapse when discretising the data if there are no values within a given range. To overcome this problem, boxplots were generated for each feature and examined manually. Custom labels and percentile levels were selected for each feature based on the shape of the boxplot. This also allowed manual selection of the number of groups into which the data were divided on a per-feature basis. Full details of these groupings are given in the supplementary material. For categorical features, the categories were used as bins.

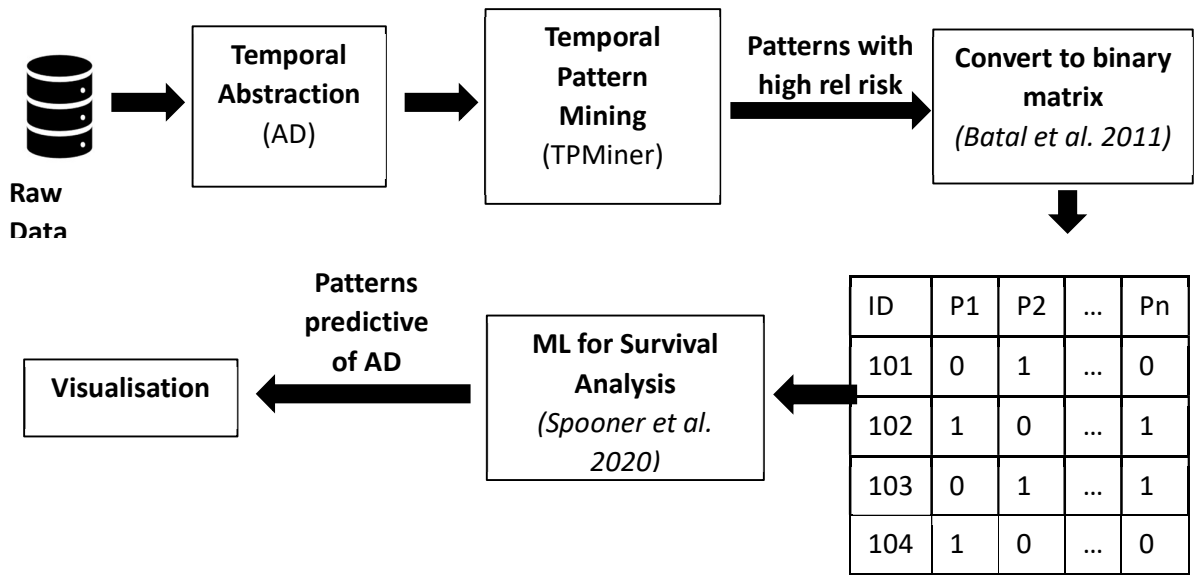

*Figure 1 The Temporal Pattern Mining Framework.*

Following abstraction of the raw values into higher-level states, consecutive states with the same label and value were aggregated to form longer patterns, ready to be mined using the temporal pattern mining algorithm. If a value was missing in a given wave, that value was assumed to remain the same as the previous wave, until a new value was recorded. This process is illustrated in Figure X which shows the BMI values for one patient over time.

| Label | Values (kg) |
|---|---|
| Underweight | Less than 18.5 |
| Normal weight | 18.5 – 24.9 |
| Overweight | 25 - 29.9 |
| Obese | 30 and above |

*Table 2 - Temporal abstraction of the Body Mass Index (BMI) using cut-off values from* https://www.cdc.gov/healthyweight/assessing/index.html

| Label | Percentile |
|---|---|
| Very low (VL) | Below 5th percentile |
| Low (L) | 5th – 25th percentile |

| Normal (N) | 25th – 75th |
| High (H) | 75th – 95th |
| Very High (VH) | Above 95th |

Table 3 - Temporal abstraction of continuous numeric values using percentiles.

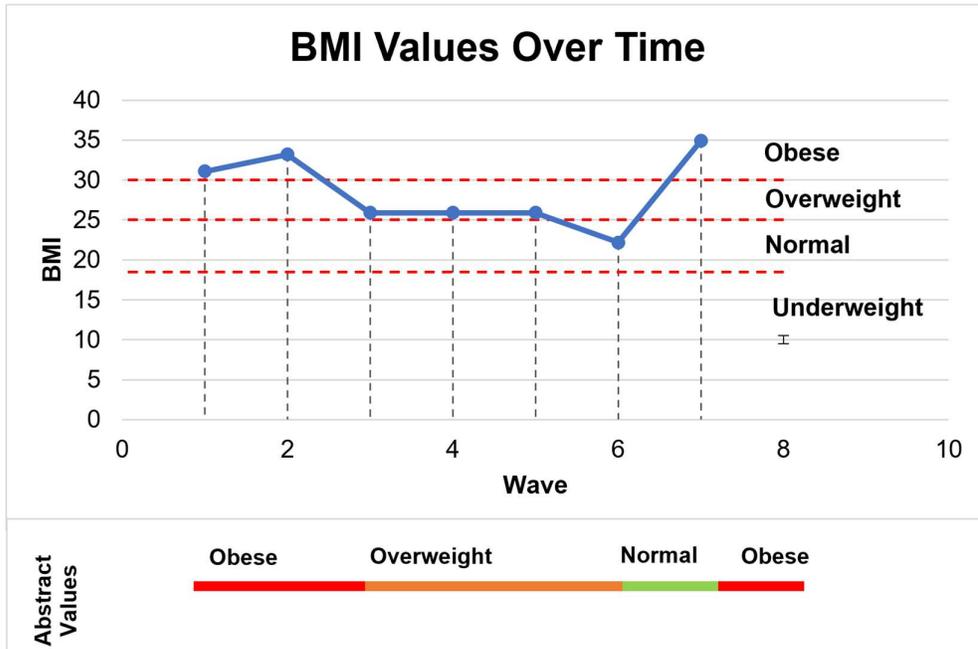

*Figure 2. An example showing changes in the BMI value over time for one study participant, illustrating how the abstracted values are formed. This illustration follows the format used by Batal et al [5].*

### 3.2.2 Temporal Pattern Mining

The proposed temporal pattern mining algorithm, C-TPM, is based on the TPMiner algorithm [7], an efficient, scalable algorithm which is itself based on the PrefixSpan algorithm [15]. The reader is referred to the relevant papers for full details of these algorithms. Both use a projection-based pattern-growth approach, whereby a database is recursively projected into a set of smaller projected databases, and patterns are grown in each projected database by calculating support only in the smaller database [15]. C-TPM follows this model and the pseudo-code for the algorithm is shown in Algorithm 1.

Each patient's data are converted into a sequence of endpoints. A prefix is a subsequence of any length that begins the sequence. A suffix is the remainder of the sequence, following the prefix. The projected database with respect to prefix α is the collection of suffixes that have α as their prefix. For efficiency, C-TPM uses a technique called pseudo-projection [15]. Rather than copying the entire set of suffixes for each prefix in each projected database, C-TPM stores the endpoint sequences only once and keeps track of the location of the suffixes within the endpoint sequence for each prefix and projected database.

**Algorithm 1:** *C-TPM(db, minsup, risk_sup)*
**Input: db**: a temporal database, **minsup**: the minimum support threshold, **rr_sup**: the minimum threshold for relative risk
**Output: TP**: Set of all temporal patterns, **proj**: Set of projected databases
1. TP <- ∅
2. Transform data into endpoint representation, ignoring "normal" levels
3. Count the support of each endpoint in the abnormal group and in the whole population
4. *risk* <- relative risk (or odds ratio) of each endpoint in the database
5. *fe* <- all endpoints with support > minsup and risk > risk_sup **//risk pruning**
6. **for** each **starting** endpoint *alpha* in *fe* **do**
7.    construct the projected database for *db_alpha*
8.    **TPSpan**(*alpha, db_alpha, min_sup, risk_sup, TP*)
9. **return** TP

function *TPSpan* (*alpha, db_alpha, minsup, risk_sup, TP*)
1. *sup, sup_a* <- **count_support**(*alpha, db_alpha*) **// scan-pruning**
2. *risk* <- relative risk (or odds ratio) of each endpoint in the projected database
3. *fe* <- *endpoints* with support > minsup and risk > *risk_sup* **// risk pruning**
4. *fe* <- **point_pruning**(*fe, alpha*) **// point pruning**
5. **for** each endpoint *ep* in *fe* **do**
6.    append *ep* to *alpha* to form *alpha_new*
7.    **if** *risk(alpha_new) > risk(alpha)* **then** **// risk pruning**
8.      *db_new* <- **construct_db**(*alpha_new, ep, risk*)
9.      **if** *alpha_new* has not been seen before **then** **// duplicate pruning**
10.        **if** all endpoints in *alpha_new* appear in pairs **then**
11.          TP <- TP U *alpha_new*
12.          **TPSPan**(*alpha_new, db_new, minsup, risk_sup, TP*)

function *count_support(alpha, db_alpha)*
1. **for** each patient's endpoint sequence **do**
2.    find the first finishing endpoint which has a corresponding starting endpoint in prefix *alpha* and mark this position
3.    count the support of every endpoint from the start of the sequence to this position, in both the whole population and the abnormal group
4. **return** support in whole population, support in abnormal group

function *construct_db(index, alpha, risk)*
1. *idx* <- index of new projected db
2. find the position of all suffixes of prefix *alpha* for each patient
3. **for** each suffix *s* **do**
     eliminate the finishing endpoints in *s* that have no starting endpoint in *alpha*
4. **return** *idx*

function *point_pruning(fe, alpha)*
1. *temp_points* <- ∅
2. **for** each endpoint *ep* in *fe* **do**
3.    **if** *ep* is a finishing endpoint **then**
4.      **if** there is a corresponding starting endpoint in *alpha* then
5.        *temp_points* <- *temp_points* U *ep*
6.    **if** *ep* is a starting endpoint **then**
7.      *temp_points* <- *temp_points* U ep
8. **return** *temp_points*

A key component of any pattern mining algorithm is its pruning strategies, which reduce the number of patterns to be searched without neglecting any important patterns, and therefore increase computational efficiency. The lack of effective pruning strategies could cause an exponential explosion in the number of patterns to be searched, making the method computationally infeasible, consuming a large amount of memory and finding many spurious patterns.

C-TPM adopts the three pruning strategies used by TPMiner and also adds some additional pruning strategies of its own. TPMiner's three pruning strategies are scan-pruning, point-pruning and postfix-pruning, all of which stem from the fact that interval endpoints must occur in pairs.

Scan-pruning recognises that it is unnecessary to scan each sequence from beginning to end. A sequence need only be scanned from the beginning to the first finishing endpoint that has a corresponding starting endpoint in prefix α. This is because endpoints occur in pairs and a sequence will never become a pattern if it doesn't contain matching pairs of endpoints. Similarly, point-pruning involves only projecting the finishing endpoints which have corresponding starting endpoints in their prefix, and postfix-pruning eliminates finishing endpoints in the suffix that have no starting endpoint in the prefix.

C-TPM adds three more pruning strategies that aid it in mining clinical data. First, as the method is searching for patterns with high relative risk, an endpoint is only added to a pattern if the new pattern increases the relative risk and exceeds a relative risk threshold and a minimum support threshold, which are selected in advance by the user. This method is named risk-pruning. Normal biological functioning will not explain the development of AD. Therefore, the method is searching for patterns that deviate from normal function, and so it is reasonable to prune any normal levels of the features. Not only are normal levels likely to occur in a large percentage of the population, but they will not give us any meaningful information about disease progression. This is performed at the start of the algorithm, when the data are transformed into endpoint representation, and is named normal-pruning.

Batal et al. [5] mined clinical data from electronic health records, where events can occur asynchronously and endpoints only infrequently occur at the same time. In contrast, the data mined in this study occurs in waves, with a value recorded in general for each feature in each wave, excluding missing data. Therefore, endpoints frequently occur simultaneously. Remembering that the order of simultaneous events is irrelevant, the algorithm will detect patterns that are identical, such as the following where A and B are endpoints, + indicates a starting endpoint, - indicates a finishing endpoint and endpoints appearing in brackets occur simultaneously:

- A+, (B+, A-), B-
- A+, (A-, B+), B-

So C-TPM prunes identical patterns as soon as they become apparent. This is named duplicate-pruning.

### 3.2.3 Multiprocessing
In order to improve the efficiency of C-TPM and enable it to handle larger numbers of features, it is implemented with multi-processing. Using the Python library *multiprocess*, a fixed-size pool of processes is created upon initialisation of the algorithm. This size is specified by the user in advance.

A new process is spawned for each frequent endpoint found at the highest level of recursion. Within each new process, a projected database is created for that endpoint and all further projected databases are created recursively within that process. The library takes care of the allocation and re-allocation of these processes as needed, so no additional load balancing is implemented.

### 3.2.4 Survival Analysis

Following the method of Batal et al. [5], the temporal patterns found are transformed into a binary matrix, with one row per patient and one column per pattern. A '1' in a particular row and column indicates that that pattern was found in the data for that patient, a '0' indicates otherwise. In this case, as the data are censored, the binary matrix is analysed by a machine learning survival analysis model [31] to determine the patterns that are predictive of AD.

### 3.2.5 Pattern Visualisation

In order to assist in understanding the results produced by C-TPM, a visualisation module produces a graphical representation of the patterns with the highest relative risk. Examples of the output of this module are shown in Figures 7 and 8. The relative risk of each pattern is shown on the left, below the pattern identifier. On the right the items making up the pattern are arranged in their order of occurrence. Items that follow one another horizontally occur consecutively. Items that align vertically, start and/or end at the same time. Each item is given a colour that represents the reference level of its value. Red represents a very low value, orange a low value, green a normal value, light blue a high value and dark blue a very high value, compared to normal levels for that variable.

## 4. Results and Discussion

### 4.1 Experimental Setup

Temporal Abstraction and C-TPM were applied to two different AD datasets to search for patterns predictive of AD. All available data were used: 7 waves from the MAS dataset and 3 waves from the OATS dataset. Details of the data in these datasets are shown in Table 1.

The algorithm was tested with a range of values for the relative risk threshold and the *minsup* threshold. Ten different machine learning survival analysis algorithms were then trained on the resulting binary matrices and several performance metrics were examined, including the predictive accuracy of the survival models, the run-time of the temporal pattern mining algorithm and the number of patterns generated.

C-TPM was written in Python v3.6.7 [32]. The machine learning survival analysis models were written in R v3.6.2 [31]. The survival analysis was run within a 5-fold cross validation framework. The visual display of the discovered patterns was implemented in VPython v7 and displayed using Web VPython v3.2. Experiments were performed on the Katana computational cluster, supported by Research Technology Services at UNSW Sydney [33].

### 4.2 Knowledge Discovery

Ideally, a temporal pattern mining algorithm should return a manageable set of patterns that are easy to understand and are predictive of the disease under investigation if it is to be useful in knowledge discovery [5]. If too many patterns are discovered, many are likely to be spurious. If too few are discovered, then some important patterns may be missed.

The number of temporal patterns discovered by C-TPM using different relative risk and *minsup* thresholds is shown in Figure 2 for the MAS dataset and Figure 3 for the OATS dataset. The values of 1.5 and 2 for the relative risk threshold represent the end points of an optimal range. A relative risk threshold less than 1.5 found too many patterns and the execution time became unmanageable. In contrast, a value greater than 2 found too few patterns and often no patterns at all.

As *minsup* increases (moving from the right to the left of the graph), the number of patterns discovered decreases, as expected. Increasing the relative risk threshold also decreases the number of patterns discovered. For the MAS dataset, with a relative risk threshold of 2.0, only 3 patterns are discovered once *minsup* reaches a value of 3%.

### 4.3 Predictive Accuracy

Eight different machine learning survival analysis models were trained on each of the binary matrices produced by the temporal pattern mining algorithm with different values of *minsup* and the relative risk threshold, to test how well the C-TPM framework was able to represent and discover temporal patterns that are predictive of AD. The predictive accuracy of these models was assessed using the Concordance Index (C-Index). The C-Index measures the proportion of pairs where the observation with the higher actual survival time has the higher probability of survival as predicted by the model [34].

The survival analysis algorithms tested were penalised regression (the Lasso, the ElasticNet and the Ridge), the Cox model with gradient boosting (GLMBoost), the Cox model with likelihood-based boosting (CoxBoost), extreme gradient boosting with linear base models (XGBLinear) and the random survival forest (RSF). Further details and explanations of these algorithms can be found [31]. The performance of these algorithms is shown in Figure 6 for MAS and Figure 7 for OATS.

For the MAS dataset, the models with a relative risk threshold of 1.5 all perform as well as or better than the model with a relative risk threshold of 2 and the same *minsup*. The lower relative risk threshold will find more patterns, so clearly this is beneficial to the model performance. In most cases, lower values of the relative risk threshold and *minsup* threshold produce the highest predictive accuracy. The best performance was achieved by the XGBLinear model with a relative risk threshold of 1.5 and a *minsup* of 1%, having a C-Index of 0.8.

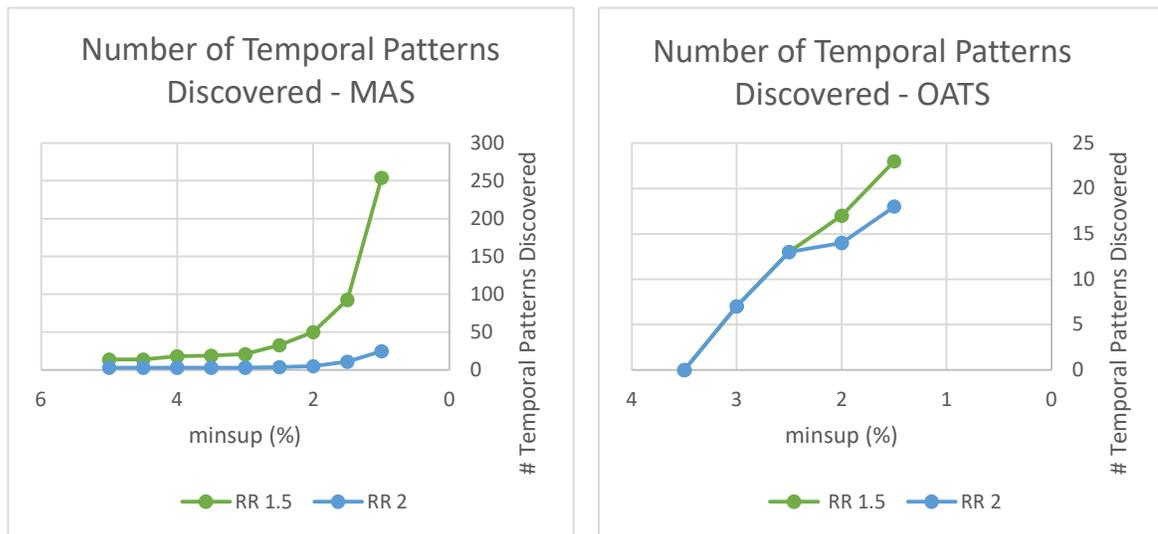

*Figure 3. Number of temporal patterns discovered by C-TPM in the MAS dataset for varying values of minsup and the relative risk threshold RR.*

*Figure 4. Number of temporal patterns discovered by C-TPM in the OATS dataset for varying values of minsup and the relative risk threshold RR.*

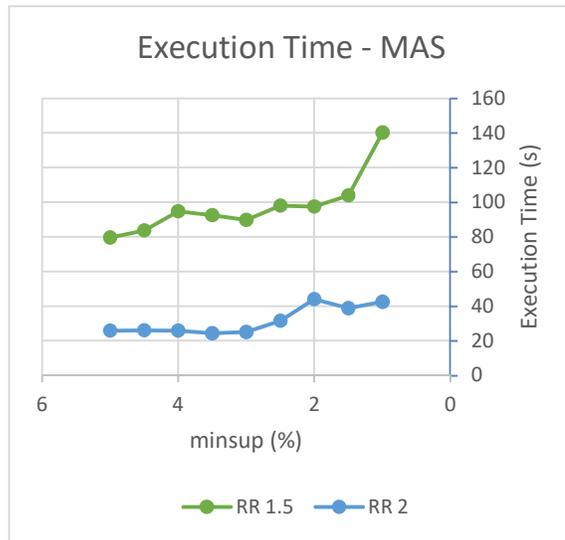
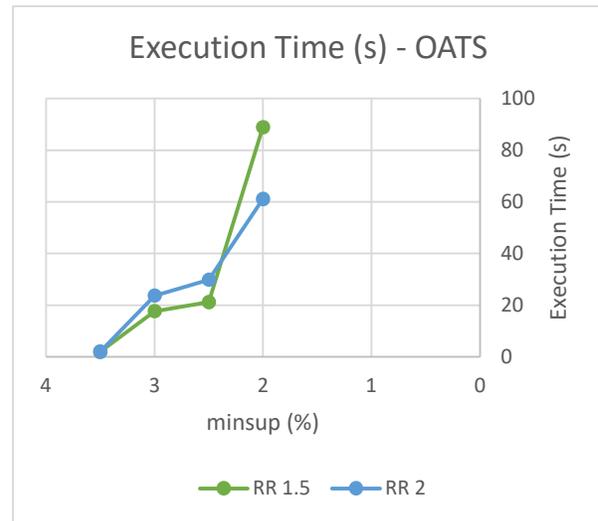

*Figure 5. Execution time in seconds for C-TPM applied to the MAS dataset with varying values of minsup and the relative risk threshold RR.*

*Figure65. Execution time in seconds for C-TPM applied to the OATS dataset with varying values of minsup and the relative risk threshold RR.*

This pattern is not repeated in the OATS dataset, where the best predictive accuracy is achieved with relatively higher values of relative risk threshold and *minsup* threshold and in many cases the models with a relative risk threshold of 2 perform as well as or better than the equivalent models with a relative risk threshold of 1.5. The best performance in the OATS dataset is also achieved by the XGBLinear model, but this time with a relative risk threshold of 2 and a *minsup* of 3%, producing a C-Index of 0.68. However, the same model with a relative risk threshold of 1.5 achieves a similar result, with a C-Index of 0.67.

The models trained on the OATS dataset exhibit a lower predictive accuracy than those trained on the MAS dataset. This behaviour is expected for several reasons: the OATS dataset has fewer samples, fewer events (diagnoses of AD) and fewer waves of data than the MAS dataset. Having more waves of data means that the patterns discovered can be longer and there is generally more information available about each patient, so it is natural that this would provide greater accuracy.

### 4.4 Execution Time

The execution times of each run of C-TPM were recorded to determine the efficiency of the algorithm. These results are shown in Figure 4 for the MAS dataset and Figure 5 for the OATS dataset. The execution times are the time spent performing pattern mining, after pre-processing steps such as loading the data and temporal abstraction.

The results of course show that the models with a higher relative risk threshold have a shorter execution time. This is a reflection of the fact that they are examining fewer endpoints and finding fewer patterns. Comparing models with the same relative risk threshold, there is an increase in execution time as the *minsup* threshold decreases, as would be expected, because a lower *minsup* value means that the method is processing more endpoints.

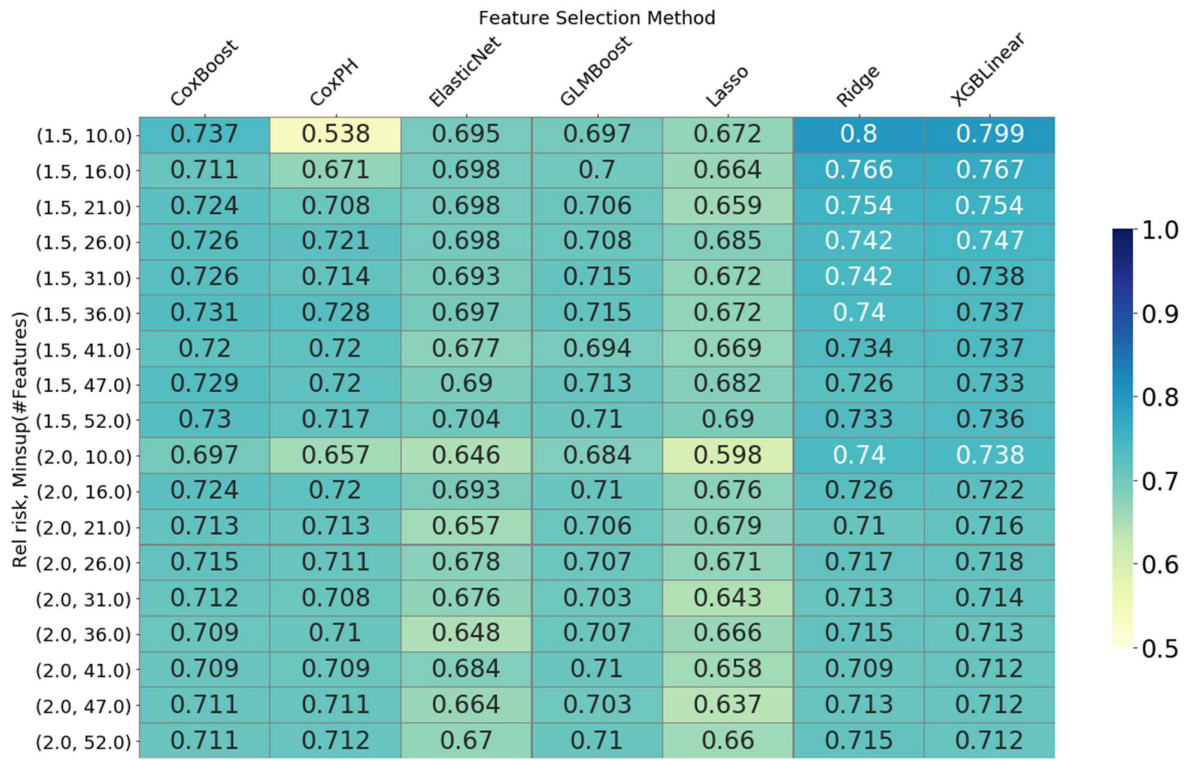

*Figure 6. Performance of the machine learning algorithms, measured by the C- Index, applied to the MAS dataset.*

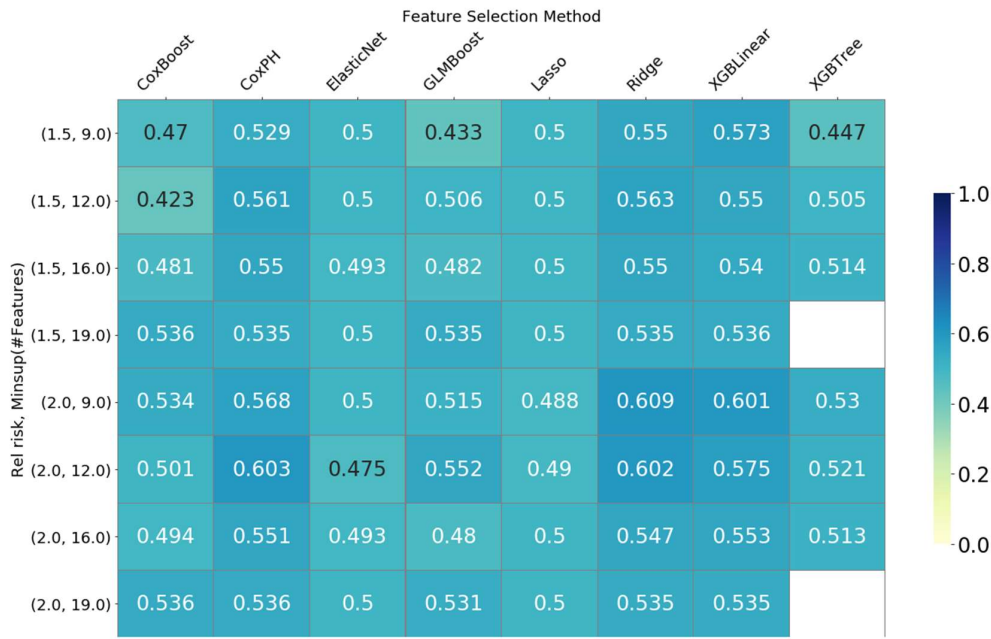

*Figure 7. Performance of the machine learning algorithms, measured by the C- Index, applied to the MAS dataset.*

For the MAS dataset, the execution times remain below or around 100 seconds until the *minsup* falls to 1%. At this point the number of patterns discovered rises significantly, as shown in Figure 2, and so the execution time also rises. However, it only rises to 140s, while modelling 70 features and finding 254 patterns. This is a significant improvement over the method of Batal et al. [5] which modelled only 5 features and recorded execution times ranging from approximately 90s at 25% *minsup* to 900s at 5% *minsup*. Of course, a direct comparison can only be made when both methods are executed on the same hardware under the same conditions, but this is not possible as Batal's method is not capable of modelling the large numbers of features that C-TPM can model.

### 4.5    Visualisation

One of the main advantages of using pattern mining to analyse longitudinal clinical data is that it can improve the interpretability of the results.  The visualisation module developed here provides a clear picture of the patterns selected by the machine learning models as predictive of AD.

There was a high degree of overlap in the patterns selected by the different machine learning models, indicating that they are relatively stable. The final set of patterns was determined from a combination of the best models. For the MAS dataset, the combination of a relative risk threshold of 1.5 and a *minsup* of 1% (10 features) produced the most accurate models. The patterns generated by each machine learning algorithm with those parameters were each ranked by their feature importance scores or coefficients, , from best to worst, and the sum of the rankings for each pattern was calculated. The patterns were then ranked again by the sum of their individual rankings. This has the effect of selecting the patterns that were ranked highly by many different models.

The top-ranking patterns centre around a small number of features, and these are explained in Table 4 for MAS and Table 5 for OATS. Interpretation of the full range of values for each of these features and other features used in the models can be found in the supplementary material.  The top 10 ranked patterns are shown in Figure 8 for MAS and Figure 9 for OATS, where only 9 patterns were selected. The highest relative risk recorded by any of the selected patterns is 5.07, meaning that patients exhibiting that pattern are 5.07 times more likely to develop AD.

Another measure of success is whether those patterns are clinically relevant. Each of the features listed in Tables 4 and 5 has been identified in the literature as a risk factor for AD and those references are included in the last column of these tables.

### 5.    Conclusion

The task of modelling multivariate, longitudinal data is a complex one, especially when applied to clinical data. Clinical data are often censored, heterogeneous and contain missing values. In addition, data from cohort studies are often collected at infrequent and slightly irregular intervals, thus ruling out the use of many analysis methods that require frequent, regularly spaced data points. However, the ability to model these data is essential in understanding the processes leading to a disease, such as AD.

This work presents a novel framework based on temporal pattern mining that can be used to successfully analyse complex clinical data and find patterns that accurately predict the onset of AD. The survival analysis models developed here achieve a C-Index of up to 0.8, showing a high degree of predictive accuracy and the patterns discovered can be clearly visualised to aid in interpretation.

The strength of this method lies in its flexibility. These techniques may be applied to any temporal data, including clinical data from other medical domains or non-clinical data from other domains by supplying appropriate cut-off values for the temporal abstraction module. The survival analysis models can be replaced by classification or regression models if that is more appropriate. The method can handle irregularly sampled temporal data, data containing missing values, heterogeneous data and censored data.

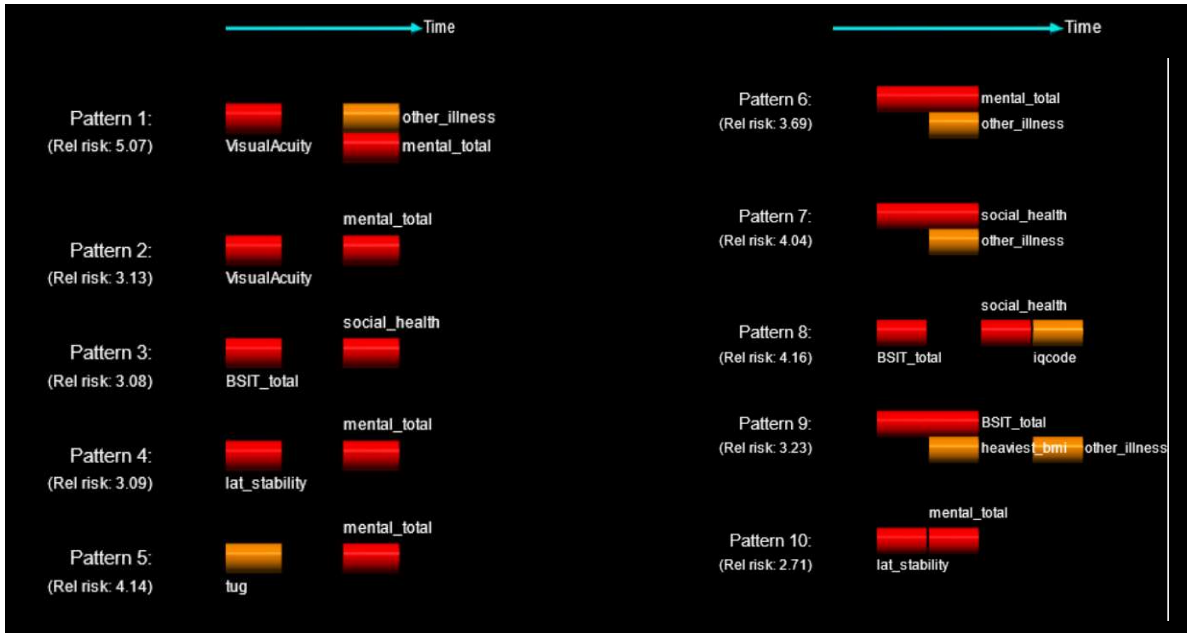

*Figure 8. Visualisation of the top patterns in the MAS dataset discovered by C-TPM. Each labelled cylinder represents one feature and its state, with the endpoints of the cylinder showing the relative starting and finishing times of that state. The particular state of the feature can be determined by its colour - orange implies a moderate level of deterioration, and red a severe level of deterioration from the normal. A pattern is made up of one or more features and their states over time, with the progression of time shown by the blue arrow at the top of the figure.*

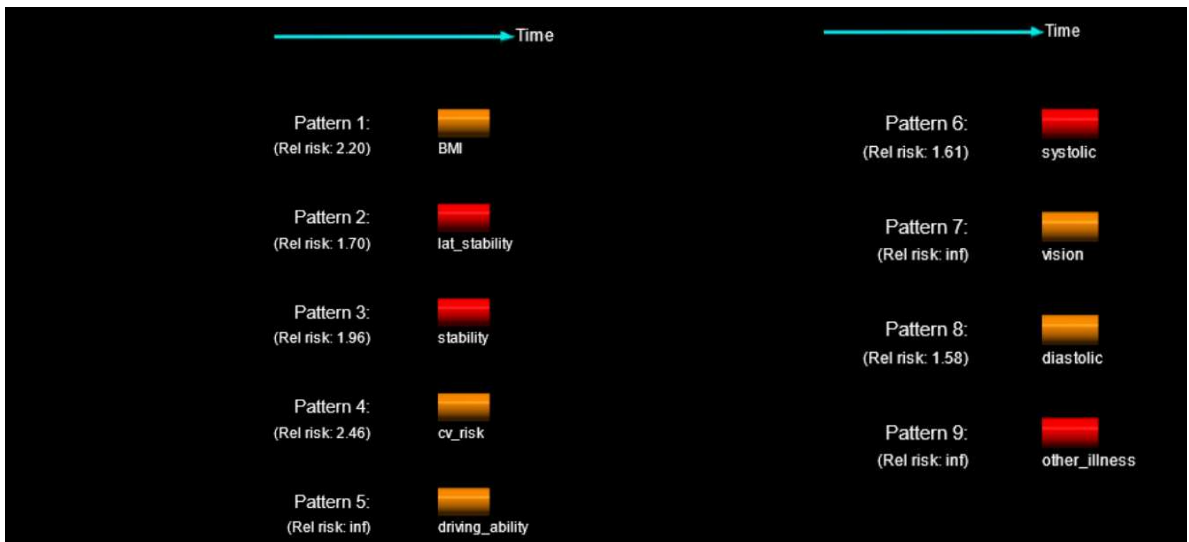

*Figure 9. Visualisation of the top patterns in the OATS dataset discovered by C-TPM. Each labelled cylinder represents one feature and its state, with the endpoints of the cylinder showing the relative starting and finishing times of that state. The particular state of the feature can be determined by its colour - orange implies a moderate level of deterioration, and red a severe level of deterioration from the normal. A pattern is made up of one or more features and their states over time, with the progression of time shown by the blue arrow at the top of the figure.*

| Abbreviation | Description | Interpretation | Ref |
|---|---|---|---|
| BSIT_total | Total score from the Brief Smell Identification Test | Abnormal: < 9 | [35] |
| cvdrisk | Risk of cardio- vascular disease – Framingham score | Low risk: < 10% | [36] |
| heaviest_bmi | Largest BMI during lifetime | Overweight: 25.0 – 29.9 | [37] |
| iqcode | Total score from the Informant Questionnaire on Cognitive Decline in the Elderly | A b[38]it worse: 4 | |
| lat_stability | Lateral stability test (30 seconds) | Poor: <= 6 secs or below 25th percentile | [39] |
| mental_total | A composite score comprising 1 point for each activity: paid work, voluntary work, television, radio, newspapers, magazines, books, classical music, games, museums, Internet, artistic pastime, 2nd language, new sport, new pastime, new evening class, new certificate course, new degree | Very low: <= 4 or below 5th percentile | [40] |
| other_illness | Number of other serious illnesses, apart from AD | Moderate: 1 other illness | [38] |
| social_health | A composite score comprising 1 point for each of: Married, not living alone, member of a group, has someone to confide in, #face to face contacts, #regular contacts | Very low: <= 3 or below 5th percentile | [41] |
| tug | Timed up and go score | Abnormal: >12 secs | [42] |
| VisualAcuity | Visual Acuity – larger of left and right eye | Moderate: moderate amount of visual loss (<6/18 - 6/60) | [43] |

*Table 4. - Explanation of the features and their values that form the top temporal patterns selected by a combination of models from the MAS dataset.*

| Abbreviation | Description | Interpretation | Ref |
|---|---|---|---|
| BMI | Body Mass Index | Overweight: 25 – 29.9 | [37] |
| cv_risk | Composite variable encoding any diagnoses of hypertension, high cholesterol, diabetes and claudication. | Medium risk: at least one of the 4 conditions diagnosed. | [36] |
| diastolic | Diastolic blood pressure | Hypertension: >= 90 Hg | [36] |
| driving_ability | Composite variable encoding the number of major and minor at fault motor vehicle accidents in the past 18 months and any driving restrictions. | Fair: Between the 25[th] and 75[th] percentile | [44] |
| lat_stability | Lateral stability test (30 seconds) | Poor: <= 6 secs or below 25th percentile | [39] |
| other_illness | Number of other serious illnesses, apart from AD | Moderate: 1 other illness | [38] |
| stability | Composite variable encoding dizziness, balance assessment and # falls in the last 18 months. | High: greater than 75[th] percentile. | [39] |
| systolic | Systolic blood pressure – average of 3 measurements | Hypertension: >= 140 Hg | [36] |
| vision | Is vision adequate for all purposes with glasses on? | Fair: problematic vision | [43] |

*Table 5. - Explanation of the features and their values that form the top temporal patterns selected by a combination of models from the OATS dataset.*

## Bibliography

[1]   P. Senin, "Dynamic Time Warping Algorithm Review," *Science (80-. ).*, vol. 2007, no. December, pp. 1–23, 2008.

[2]   D. E. Rumelhart, G. E. Hinton, and R. J. Williams, "Learning Internal Representations by Error Propagation," *Tech. rep. ICS 8504. San Diego, Calif. Inst. Cogn. Sci. Univ. Calif.*, 1985.

[3]   J. Li *et al.*, "Mining risk patterns in medical data," *Proc. ACM SIGKDD Int. Conf. Knowl. Discov. Data Min.*, pp. 770–775, 2005.

[4]   J. Allen, "Maintaining knowledge about temporal intervals," *Commun. ACM*, vol. 26, no. 11, pp. 832–843, 1983.

[5]   I. Batal, H. Valizadegan, and G. Cooper, "A Pattern Mining Approach for Classifying Multivariate Temporal Data," *Proc. IEEE Int Conf Bioinforma. Biomed*, no. November 12, pp. 358–365, 2011.

[6]   Y. Shahar and M. A. Musen, "Knowledge-based temporal abstraction in clinical domains," *Artif. Intell. Med.*, vol. 8, pp. 267–298, 1996.

[7]   Y. C. Chen, W. C. Peng, and S. Y. Lee, "Mining temporal patterns in interval-based data," *2016 IEEE 32nd Int. Conf. Data Eng. ICDE 2016*, vol. 27, no. 12, pp. 1506–1507, 2016.

[8]   "Alzheimer's Association: Facts and Figures." [Online]. Available: https://www.alz.org/alzheimers-dementia/facts-figures. [Accessed: 10-Dec-2021].

[9]   D. J. Selkoe and J. Hardy, "The amyloid hypothesis of Alzheimer's disease at 25 years.," *EMBO Mol. Med.*, vol. 8, no. e201606210, pp. 1–14, 2016.

[10]  A. Kocheturov, P. Momcilovic, A. Bihorac, and P. M. Pardalos, "Extended vertical lists for temporal pattern mining from multivariate time series," *Expert Syst.*, vol. 36, no. 5, pp. 1–16, 2019.

[11]  R. Agrawal and R. Srikant, "Mining sequential patterns," *Proc. - Int. Conf. Data Eng.*, pp. 3–14, 1995.

[12]  J. Pei *et al.*, "PrefixSpan: Mining sequential patterns efficiently by prefix-projected pattern growth," *Proc. - Int. Conf. Data Eng.*, pp. 215–224, 2001.

[13]  M. J. Zaki, "Scalable Algorithms for Association Mining," *IEEE Trans Knowl Data Eng*, vol. 12, no. 3, pp. 372–390, 2000.

[14]  J. Han, J. Pei, Y. Yin, and R. Mao, "Mining frequent patterns without candidate generation - a Frequent-Pattern Tree Approach," *Data Min. Knowl. Discov.*, vol. 8, no. 2, pp. 53–87, 2004.

[15]  J. Pei *et al.*, "Mining sequential patterns by pattern-growth: The prefixspan approach," *IEEE Trans. Knowl. Data Eng.*, vol. 16, no. 11, pp. 1424–1440, 2004.

[16]  E. Winarko and J. F. Roddick, "ARMADA - An algorithm for discovering richer relative temporal association rules from interval-based data," *Data Knowl. Eng.*, vol. 63, no. 1, pp. 76–90, 2007.

[17]  Y. C. Chen, J. C. Jiang, W. C. Peng, and S. Y. Lee, "An efficient algorithm for mining time interval-based patterns in large databases," *Int. Conf. Inf. Knowl. Manag. Proc.*, pp. 49–58, 2010.

[18]  S. Y. Wu and Y. L. Chen, "Mining nonambiguous temporal patterns for interval-based events," *IEEE Trans. Knowl. Data Eng.*, vol. 19, no. 6, pp. 742–758, 2007.

[19]  D. Patel, W. Hsu, and M. L. Lee, "Mining relationships among interval-based events for classification," *Proc. ACM SIGMOD Int. Conf. Manag. Data*, pp. 393–404, 2008.

[20]  R. Moskovitch and Y. Shahar, "Classification of multivariate time series via temporal abstraction and time intervals mining," *Knowl. Inf. Syst.*, vol. 45, no. 1, pp. 35–74, 2015.

[21]  I. Batal, H. Valizadegan, G. F. Cooper, and M. Hauskrecht, "A temporal pattern mining approach for classifying electronic health record data," *ACM Trans. Intell. Syst. Technol.*, vol. 4, no. 4, pp. 1–22, 2013.

[22] L. Geng and H. J. Hamilton, "Interestingness measures for data mining: A survey," *ACM Comput. Surv.*, vol. 38, no. 3, p. 3, 2006.

[23] H. Li, J. Li, L. Wong, M. Feng, and Y. P. Tan, "Relative risk and odds ratio: A data mining perspective," *Proc. ACM SIGACT-SIGMOD-SIGART Symp. Princ. Database Syst.*, pp. 368–377, 2005.

[24] J. Li, A. W. chee Fu, and P. Fahey, "Efficient discovery of risk patterns in medical data," *Artif. Intell. Med.*, vol. 45, no. 1, pp. 77–89, 2009.

[25] S. Tenny and M. R. Hoffman, "Relative Risk," *StatPearls [Internet]. Treasure Isl. StatPearls Publ.*, 2021.

[26] K. Orphanou, A. Dagliati, L. Sacchi, A. Stassopoulou, E. Keravnou, and R. Bellazzi, "Combining Naive Bayes Classifiers with Temporal Association Rules for Coronary Heart Disease Diagnosis," *Proc. - 2016 IEEE Int. Conf. Healthc. Informatics, ICHI 2016*, pp. 81–92, 2016.

[27] D. M. Maslove, T. Podchiyska, and H. J. Lowe, "Discretization of continuous features in clinical datasets," *J. Am. Med. Informatics Assoc.*, vol. 20, no. 3, pp. 544–553, 2013.

[28] P. S. Sachdev *et al.*, "The Sydney Memory and Ageing Study (MAS): methodology and baseline medical and neuropsychiatric characteristics of an elderly epidemiological non-demented cohort of Australians aged 70-90 years.," *Int. Psychogeriatr.*, vol. 22, no. 8, pp. 1248–1264, 2010.

[29] P. S. Sachdev *et al.*, "A comprehensive neuropsychiatric study of elderly twins: the Older Australian Twins Study.," *Twin Res. Hum. Genet.*, vol. 12, no. 6, pp. 573–82, 2009.

[30] Y. Shahar, "Dimension of time in illness: An objective view," *Ann. Intern. Med.*, vol. 132, no. 1, pp. 45–53, 2000.

[31] A. Spooner, A. Sowmya, P. Sachdev, N. A. Kochan, J. Trollor, and H. Brodaty, "Machine Learning Models for Predicting Dementia – a Comparison of Methods for Survival Analysis of High-Dimensional Clinical Data," *Nat. Sci. Reports*, pp. 1–10, 2020.

[32] "The Python Language Reference." [Online]. Available: https://docs.python.org/3/reference/.

[33] "Katana Computational Cluster," *https://dx.oi.org/10.26190/669x-a286*.

[34] F. E. Harrell, R. M. Califf, D. B. Pryor, K. L. Lee, and R. A. Rosati, "Evaluating the Yield of Medical Tests," *JAMA J. Am. Med. Assoc.*, vol. 247, no. 18, pp. 2543–2546, 1982.

[35] S. L. Risacher *et al.*, "Olfactory identification in subjective cognitive decline and mild cognitive impairment: Association with tau but not amyloid positron emission tomography," *Alzheimer's Dement. Diagnosis, Assess. Dis. Monit.*, vol. 9, pp. 57–66, 2017.

[36] T. C. De Toledo Ferraz Alves, L. K. Ferreira, M. Wajngarten, and G. F. Busatto, "Cardiac disorders as risk factors for Alzheimer's disease," *J. Alzheimer's Dis.*, vol. 20, no. 3, pp. 749–763, 2010.

[37] G. Livingston *et al.*, "Dementia prevention, intervention, and care: 2020 report of the Lancet Commission," *Lancet*, vol. 396, no. 10248, pp. 413–446, 2020.

[38] D. M. Lipnicki *et al.*, "Risk Factors for Late-Life Cognitive Decline and Variation with Age and Sex in the Sydney Memory and Ageing Study," *PLoS One*, vol. 8, no. 6, 2013.

[39] N. Mesbah, M. Perry, K. D. Hill, M. Kaur, and L. Hale, "Postural stability in older adults with alzheimer disease," *Phys. Ther.*, vol. 97, no. 3, pp. 290–309, 2017.

[40] Scott Gottlieb, "Mental activity may help prevent dementia," *BMJ*, vol. 326, no. 7404, p. 1418, 2003.

[41] R. S. Wilson *et al.*, "Loneliness and risk of Alzheimer disease," *Arch. Gen. Psychiatry*, vol. 64, no. 2, pp. 234–240, 2007.

[42] J. E. Lee *et al.*, "Association between timed up and go test and future dementia onset," *Journals Gerontol. - Ser. A Biol. Sci. Med. Sci.*, vol. 73, no. 9, pp. 1238–1243, 2018.

[43] H. R. Davies-Kershaw, R. A. Hackett, D. Cadar, A. Herbert, M. Orrell, and A. Steptoe,

"Vision Impairment and Risk of Dementia: Findings from the English Longitudinal Study of Ageing," *J. Am. Geriatr. Soc.*, vol. 66, no. 9, pp. 1823–1829, 2018.

[44] S. Bayat *et al.*, "GPS driving: a digital biomarker for preclinical Alzheimer disease," *Alzheimer's Res. Ther.*, vol. 13, no. 1, pp. 1–9, 2021.